\documentclass[10pt, a4paper]{article}
\usepackage{lrec2014}
\usepackage{graphicx}

\usepackage{epstopdf}
\usepackage[latin1]{inputenc}
\usepackage{authblk}
\usepackage{url}

\newcommand{\ignore}[1]{}

\title{TabMCQ: A Dataset of General Knowledge Tables and Multiple-choice Questions}

\name{Sujay Kumar Jauhar$^{\ast}$, Peter Turney$^{\dagger}$, Eduard Hovy$^{\ast}$}

\address{$^{\ast}$Carnegie Mellon University \\ Pittsburgh, PA, USA \\ \{sjauhar, hovy\}@cs.cmu.edu \\ \\
         $^{\dagger}$ Allen Institute for Artificial Intelligence \\ Seattle, WA, USA \\ petert@allenai.org}

\abstract{We describe two new related resources that facilitate modelling of general knowledge reasoning in 4th grade science exams. The first is a collection of curated facts in the form of tables, and the second is a large set of crowd-sourced multiple-choice questions covering the facts in the tables. Through the setup of the crowd-sourced annotation task we obtain implicit alignment information between questions and tables. We envisage that the resources will be useful not only to researchers working on question answering, but also to people investigating a diverse range of other applications such as information extraction, question parsing, answer type identification, and lexical semantic modelling.  \\ \newline }

\begin{document}

\maketitleabstract

\section{Introduction}

The Aristo project at AI2 \cite{clark2015elementary} uses standardized science exams as drivers for research in Artificial Intelligence. Aristo's question answering format exposes a variety of interesting problems and challenges in NLP \cite{weston2015towards}, such as information extraction, semantic modelling and reasoning. An important component of a system that performs Question Answering (QA) is a store of background knowledge for fact-checking and reasoning. This store can be in a variety of modes and formalisms: large-scale extracted and curated knowledge bases \cite{fader2014open}, structured models such as Markov Logic Networks \cite{khotexploring}, or simple text corpora in information retrieval approaches \cite{tellex2003quantitative}.

There is, however, a fundamental trade-off in the expressive power of a formalism and its ability to be curated easily using existing data and tools. In this paper we describe our work on building knowledge tables, whose semi-structured format affords a balance between expressive power and ease of creation. Additionally, we introduce our  efforts at building a large bank of Multiple-Choice Questions (MCQs) using crowd-sourcing on Mechanical Turk (MTurk) by imposing structural constraints on the MCQs from the tables. These constraints lead to consistency in MCQ quality and additionally enable us to harvest alignment information between questions and table cells with little additional effort from annotators. The tables, the MCQs, and the alignments between the two constitute a trio of resources that are, to the best of our knowledge, a first of their kind. We are releasing this dataset publicly\footnote{\url{http://allenai.org/content/data/TabMCQ_v_1.0.zip}}, and we believe it will not only be useful to those working on question answering, but present interesting challenges to researchers exploring a number of related areas.

\section{Related Work}

In related recent work \newcite{pasupat2015compositional} create a dataset of QA pairs over tables. However, their annotation setup does not impose structural constraints from tables, and produces simple QA pairs rather than MCQs. \cite{yin2015neural} and \cite{neelakantan2015neural} use tables in the context of question answering, but deal with synthetically generated query data for those tables.  More generally tables have been related to QA in the context of queries over relational databases \cite{cafarella2008webtables,pimplikar2012answering}. Regarding crowd-sourcing for question creation, \newcite{aydin2014crowdsourcing} harvest MCQs via a gamified app. However their work does not involve tables. Monolingual alignment datasets have also been explored separately, for example by \newcite{export:70453} in the context of Textual Entailment.

\section{Motivation for the Dataset}
\label{sec:motiv}

Our motivation for constructing and releasing this data stems from promising preliminary results on using tables and their alignments to MCQs for solving AI2's Aristo challenge. We manually annotated 77 of the 108 4th grade science exam questions in the Regents dataset\footnote{\url{http://allenai.org/content/data/Regents.zip}} for alignment to tables. The resulting data was exactly like the one in this paper, but at a fraction of the scale. Even with this small amount of data we were able to build a system that rivaled our best in-house solvers of the Aristo challenge. A larger dataset will enable us to exceed this performance, and also allow us to explore models that require greater amounts of data (such as Neural Networks).

A detailed description of our system is beyond the scope of this paper, but in summary it uses a feature-rich log-linear model to score table cells on their relevance to answering a question. A combination of structural and textual features are used to capitalize on the semi-structured format of tables.

\section{General Knowledge Tables}

In this section we present the set of tables of curated natural language facts. We discuss their semi-structured form and its utility for knowledge representation. We then detail the data and possibilities for its extension.

\subsection{Tables as a Form of Knowledge Representation}
\label{subsec:know-rep}

\begin{table*}
\centering
\begin{tabular}{c|c|c|c|c|c|c}
\textbf{Phase Change} &  & \textbf{Initial State} &  & \textbf{Final State} &  & \textbf{Heat Transfer}\tabularnewline
\hline 
\hline 
Melting & causes a & solid & to change into a & liquid & by & adding heat\tabularnewline
\hline 
Vaporization & causes a & liquid & to change into a & gas & by & adding heat\tabularnewline
\hline 
Condensation & causes a & gas & to change into a & liquid & by & removing heat\tabularnewline
\hline 
Sublimation & causes a & solid & to change into a & gas & by & adding heat\tabularnewline
\end{tabular}
\caption{Example part of table concerning phase state changes.}
\label{tab:exmpl}
\end{table*}

An example of the kind of table we construct is given in Table~\ref{tab:exmpl}. This format is semi-structured: the rows of the table (with the exception of the header) are essentially a list of sentences, but with well-defined recurring filler patterns. Together with the header, these patterns divide the rows into meaningful columns.

The resulting table structure has some interesting semantics. A row in a table corresponds to a \emph{fact} in the world\footnote{Rows may also be seen as predicates, or more generally frames with typed arguments.}. The cells in a row correspond to concepts, entities, or processes that participate in this fact. A content column\footnote{We differentiate these from filler columns, which only contain a recurring pattern, and no information in their header cells.} corresponds to a group of concepts, entities, or processes that are the same \emph{type}. The head of the column is an abstract description of the \emph{type}. We may view the head as a hypernym and the cells in the column below as co-hyponyms of the head.

The structure of tables also exposes sets of analogies. More formally, let $row_i$ and $row_j$ be any two rows in a table. Let $col_k$ and $col_m$ be any two content columns in a table. The four cells where the rows and columns intersect form an analogy: $cell_{i,k}$ is to $cell_{i,m}$ as $cell_{j,k}$ is to $cell_{j,m}$. That is, the relation between $cell_{i,k}$ and $cell_{i,m}$ is highly similar to the relation between $cell_{j,k}$ and $cell_{j,m}$. For example, in Table~\ref{tab:exmpl} one such analogy is: ``Melting'' is to ``solid'' as ``Condensation'' is to ``gas''. Here the latent relation being represented is the one between a ``Phase Change'' and the ``initial state'' of a substance upon which it acts. The table is essentially a compact representation of a large number of scientific analogies.


The semi-structured nature of this data is flexible. Since facts are presented as sentences, the tables can simply act as a text corpus for information retrieval. Depending on the end application, more complex models can rely on the inherent structure in tables. They can be used as is for information extraction tasks that seek to zero in on specific nuggets of information, or they can be converted to logical expressions using rules defined over neighboring cells. Regarding construction, the recurring filler patterns can be used as templates to extend the tables semi-automatically by searching over large corpora for similar facts \cite{ravichandran2002learning}. 

\subsection{The Dataset}

Our dataset of tables was, for the most part, constructed manually by the second author of this paper and a knowledge engineer. The target domain for the tables was the Regents 4th grade science exam. The majority of tables were constructed to contain topics and facts in this exam (with additional facts being added once a coherent topic had been identified), with the rest being targeted at an additional in-house question bank of 500 questions.

The dataset consists of 65 hand-crafted tables organized topic-wise, from bounded tables such as the one about phase changes, to virtually unbounded tables, such as the kind of energy used in performing an action. An additional collection of 5 semi-automatically generated tables contains a more heterogeneous mix of facts across topics. A total of 3851 facts, or information rows, are present in the manually constructed tables, while an additional 4415 are present in the semi-automatically generated ones. The number of content columns in a given table varies from 2 to 5.

\subsection{Future Extensions}

While the tables are reasonably complete with regard to the target domain for which they were constructed, they are not comprehensive for any open-ended knowledge retrieval task. We are in the process of investigating ways to extend the tables semi-automatically to cover a wider collection of facts, by using the set of existing facts as seeds. An in-house fuzzy search engine allows us to search over patterns consisting not only of words and wild-card symbols, but parts-of-speech and semantically related concepts.

An interesting avenue for future research is to completely automate the process of this extension, and even construct semi-structured knowledge tables on the fly for targeted queries.

\section{Crowd-sourcing Multiple-choice Questions from Tables}
\label{sec:cs-mcq}

In this section we outline the MCQs generated from tables. We first describe the annotation task and argue for constraints imposed by table structure. We next describe the data we generated via crowd-sourcing. Finally, we also list some possible extensions to the kinds of questions we generate.

\subsection{The Annotation Task -- Constraining MCQs with Tables}

We use MTurk to generate MCQs by imposing constraints derived from the structure of the tables. These constraints help annotators create questions with scaffolding information, and lead to consistent quality in the generated output. An additional benefit of this format is the alignment information from cells in the tables to the MCQs that are generated as a by-product.

\begin{figure*}
\centering
\includegraphics[width=1.0\textwidth]{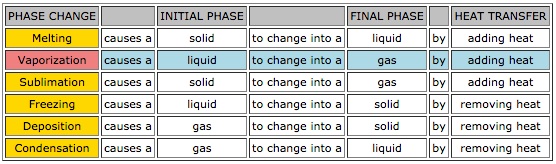}
\caption{Example table from MTurk annotation task illustrating constraints. We ask Turkers to construct questions from blue cells, such that the red cell is the correct answer, and distractors must be selected from yellow cells.}
\label{fig:mturktab}
\end{figure*}

The annotation task guides a Turker through the process of creating an MCQ. Given a table, we choose a target cell to be the correct answer for a new MCQ. First, we ask Turkers to create a question by primarily using information from the rest of the row containing the target cell, in such a way that the target cell is its correct answer. Then annotators must select all the cells in the row that they actually used to construct the question. Following this, Turkers must construct 4 succinct choices for the question, one of which is the correct answer and the other 3 are distractors. These distractors must be formed from other cells in the column containing the target cell. If there are insufficient unique cells in the column Turkers may create their own. To simplify and reinforce table constraints over rows and columns, our interface highlights table cells as shown in Figure \ref{fig:mturktab}. Turkers are allowed to rephrase and shuffle the contents of cells to arrive at the 4 succinct choices. Consequently we require Turkers to indicate which of the 4 choices that they created is the correct one for their MCQ.

In addition to an MCQ we obtain the following alignment information with no additional effort from annotators. We know which table, row, and column contain the answer, and thus also which header cells might be relevant to the question. Additionally, we know the cells of a row that were used to construct a question.

\subsection{The Dataset}

We created an individual HIT for every content bearing cell from each one of the 65 manually constructed tables. We paid annotators a reward of 10 cents per MCQ, and asked for 1 annotation per HIT for most tables. However, for an initial set of 4 tables which we used in a pilot study, we asked for 3 annotations per HIT\footnote{The goal was to observe whether there was diversity in the MCQs created for a target cell. The results were not sufficiently conclusive to warrant a threefold increase in the cost of creation.}.

In terms of qualifications, we required Turkers to have a HIT approval rating of 95\% or higher, with a minimum of at least 500 HITs approved. Additionally, we restricted the demographics of our workers to the US.

The annotators were able to create an MCQ from a table in approximately 70 seconds. They were also largely successful in their attempts. Manual inspection of the generated output also revealed that questions are of consistently good quality. They are certainly good enough for training machine learning models and many are even good enough as evaluation data for QA. A sample of generated MCQs are presented in Table \ref{tab:mcq_exmpl}.

\begin{table*}
\centering
\begin{tabular}{c|c}
What is the orbital event with the longest day and the shortest night. & What sense organ is used to detect pitch and harmony?\tabularnewline
\hline 
\hline 
A) \textbf{Summer solstice} & \textbf{A) Ears}\tabularnewline
B) Winter solstice & B) Eyes\tabularnewline
C) Spring equinox & C) Noses\tabularnewline
D) Fall equinox & D) Tongues\tabularnewline
\hline 
A graduated cylinder is used to measure what of liquids? & Steel is a/an \_\_\_\_\_ of electricity\tabularnewline
\hline 
\hline 
\textbf{A) Volume} & A) Separator\tabularnewline
B) Density & B) Isolator\tabularnewline
C) Depth & C) Insulator\tabularnewline
D) Pressure & \textbf{D) Conductor}\tabularnewline
\end{tabular}
\caption{Example MCQs generated from our pilot MTurk tasks. Correct answer choices are in bold.}
\label{tab:mcq_exmpl}
\end{table*}

Once annotations were completed we implemented some simple sanity checks to evaluate the data before approving HITs. These included things like checking if an MCQ has at least 3 choices, whether choices are repeated, etc. Through this process we rejected about 0.7\% of submitted HITs. We had to further prune our data to discard some MCQs after accepting HITs due to corrupted data, or badly constructed MCQs. A total of 160 MCQs were lost through the cleanup. Thus our dataset of MCQs covers the overwhelming majority, but not all, of the cells in the collection of tables.

In the end our complete data consists of 9091 MCQs, which is -- to the best of our knowledge -- orders of magnitude larger than any existing collection of science exam MCQs available for research. These MCQs also come with alignment information to tables, rows, columns and cells.

\subsection{Future Extensions}

In the future we hope to generate MCQs with other kinds of constraints from tables. Currently the constraints are row-dominant: the questions are based on cells from the row in which a target cell occurs. Such questions are often information look-up questions -- the simplest kind. We plan to explore column-dominant questions, which would lead to questions about abstraction or specification (based on the hypernym-hyponym relationship in columns). Other structural semantics we hope to investigate are multi-row constraints that result in comparisons, or joins on multiple tables with related columns that results in chaining or reasoning.

\section{Utility to the Research Community}

We believe that the data we have collected will be useful to people with diverse research interests in the NLP community. While the tables were designed for facts covered in 4th grade science exams, the contents are general enough to be used as background knowledge in simple domains. The additional structure presents interesting challenges to people interested in information extraction, especially when considered with the alignments to MCQs. They can be used for applications such as question parsing and answer type extraction. The structural semantics of tables (as described in Section~\ref{subsec:know-rep}) can also present a interesting challenges for those interested in lexical semantics and analogical reasoning. Jointly, the tables and MCQs can be used for QA, -- with great effect as summarized in Section~\ref{sec:motiv}

\section{Conclusion}

We have presented a dataset of tables, MCQs and alignment information between the two. Our preliminary experiments with this trio of resources even on a much smaller scale showed promising results, rivaling the best current systems on the Aristo QA challenge. We are thus led to believe the data described in this paper will be very useful to researchers working on QA. However the usefulness of the dataset potentially extends to other areas of NLP as well. The tables and MCQs individually represent resources of interest to people requiring background knowledge or working on semantic modelling and information extraction. Moreover, alignment information is expensive and time-consuming to annotate and consequently scarce; yet alignment of textual fragments is a recurring theme in NLP. Our setup allows us to harvest this resource easily at scale, thus becoming useful to people working on problems like paraphrasing, textual entailment and question parsing.

\section{Acknowledgements}

We'd like to thank the Allen Institute for Artificial Intelligence for generously funding the creation of this dataset, and permitting its release. Thanks also go to Isaac Cowhey for his effort in painstakingly building the tables. The first and third authors of this paper were supported in part by the following grants:  NSF grant IIS-1143703, NSF award IIS-1147810, DARPA grant FA87501220342.

\bibliographystyle{lrec2014}
\bibliography{xample}

\end{document}